\documentclass[nohyperref]{article}

\usepackage{layouts}
\usepackage{microtype}
\usepackage{graphicx}
\usepackage{subfigure}
\usepackage{booktabs} 
\usepackage{multirow}
\usepackage{caption}
\usepackage{hyperref}
\usepackage{float}
\usepackage{newfloat}
\usepackage{listings}
\DeclareCaptionStyle{ruled}{labelfont=normalfont,labelsep=colon,strut=off} 
\floatstyle{ruled}
\newfloat{listing}{tb}{lst}{}
\floatname{listing}{Listing}

\usepackage[accepted]{style}

\usepackage{amsmath}
\usepackage{amssymb}
\usepackage{mathtools}
\usepackage{amsthm}
\usepackage{bbm}

\DeclareMathOperator*{\argmax}{arg\,max}

\usepackage[capitalize,noabbrev]{cleveref}
\usepackage[absolute,overlay]{textpos}

\theoremstyle{plain}
\newtheorem{theorem}{Theorem}[section]

\theoremstyle{definition}
\newtheorem{definition}[theorem]{Definition}

\theoremstyle{remark}

\usepackage[textsize=tiny]{todonotes}

\icmltitlerunning{EXACT: How to Train Your Accuracy}

\begin{document}

\twocolumn[
\icmltitle{EXACT: How to Train Your Accuracy}

\icmlsetsymbol{equal}{*}

\begin{icmlauthorlist}
\icmlauthor{Ivan Karpukhin}{tinkoff}
\icmlauthor{Stanislav Dereka}{tinkoff}
\icmlauthor{Sergey Kolesnikov}{tinkoff}
\end{icmlauthorlist}

\icmlaffiliation{tinkoff}{Tinkoff}

\icmlcorrespondingauthor{Ivan Karpukhin}{i.a.karpukhin@tinkoff.ru}

\icmlkeywords{Machine Learning, Accuracy optimization, Loss function, Cross entropy}

\vskip 0.3in
]



\printAffiliationsAndNotice{} 

\begin{textblock*}{12cm}(5.5cm,1cm) 
   \LARGE \textcolor{gray!50}{Published in Pattern Recognition Letters (2024)}
\end{textblock*}

\begin{abstract}
Classification tasks are usually evaluated in terms of accuracy. However, accuracy is discontinuous and cannot be directly optimized using gradient ascent. Popular methods minimize cross-entropy, hinge loss, or other surrogate losses, which can lead to suboptimal results.

In this paper, we propose a new optimization framework by introducing stochasticity to a model's output and optimizing expected accuracy, i.e. accuracy of the stochastic model. Extensive experiments on linear models and deep image classification show that the proposed optimization method is a powerful alternative to widely used classification losses.


\end{abstract}
\setcounter{footnote}{1}

\section{Introduction}
Accuracy is one of the most used evaluation metrics in computer vision \cite{lecun1998mnist,krizhevsky2009cifar,krizhevsky2012imagenetcls}, natural language processing \cite{maas2011nlpsentiment, zhang2015textcls}, and tabular data classification \cite{arik2021tabnet}.
While accuracy naturally appears in classification tasks, it is discontinuous and difficult to optimize directly. To tackle this problem, multiple surrogate losses were proposed, including cross-entropy and hinge loss \cite{wang2022comprehensive}. 
However, as shown below, in our toy example from Figure \ref{fig:toy}, decreasing cross-entropy or hinge loss can lead to a drop in accuracy. In other words, there is no direct connection between surrogate losses and accuracy optimization.

In this work, we take a different approach to accuracy optimization. Our idea is to introduce stochasticity to the model's output and then optimize the expected accuracy, i.e. accuracy of the stochastic model, via gradient methods.
We call the proposed method EXpected ACcuracy opTimization (EXACT). It directly optimizes the accuracy of the stochastic model in contrast to surrogate losses.

Contributions of this work can be summarized as follows:
\begin{enumerate}
\item{We propose a new optimization framework for classification tasks. To the best of our knowledge, it is the first work, where the classification model's accuracy is directly optimized via gradient methods.}
\item{We provide an efficient method for evaluating the proposed loss function and its gradient. We do this by presenting a new algorithm for gradient propagation through the orthant integral of the multivariate normal PDF \footnote{\url{https://github.com/ivan-chai/exact}}.}

\item{We compare the quality of the proposed EXACT method with cross-entropy and hinge losses. According to our results, EXACT improves accuracy in multiple tabular and image classification tasks, including SVHN, CIFAR10, and CIFAR100.}
\end{enumerate}

\begin{figure}[t]
\vskip 0.1in
\centering
\includegraphics{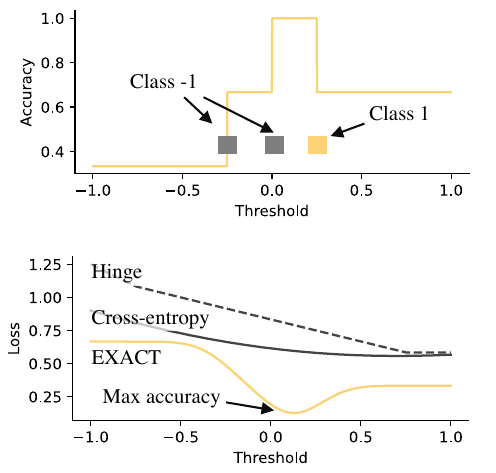}
\caption{The toy example, which demonstrates importance of accuracy optimization. The model consists of a single bias parameter (decision threshold), while scaling weight is assumed to be 1. EXACT achieves 100\% accuracy, while cross-entropy and hinge loss misclassify one element.}
\label{fig:toy}
\vskip -0.1in
\end{figure}

\begin{figure*}[ht]
\vskip 0.0in
\begin{center}
\centerline{\includegraphics[width=0.9\textwidth]{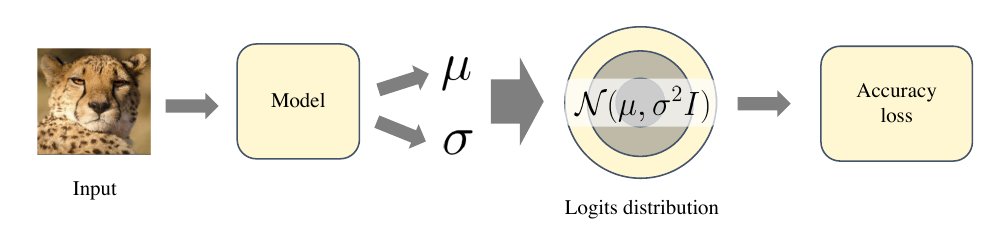}}
\caption{EXACT training pipeline. The model predicts the mean and variance of the logit vector. EXACT's training objective estimates accuracy, which is differentiable for the stochastic model.}
\label{fig:pipeline}
\end{center}
\vskip -0.2in
\end{figure*}

\section{Related Work}

\subsection{Classification Losses}
One of the most used classification loss functions is cross-entropy (CE), also known as negative log-likelihood \cite{wang2022comprehensive}. Minimization of cross-entropy reduces the difference between predicted class probabilities and true posteriors. If a model predicts true posteriors, then selecting the class with maximum probability will lead to maximum accuracy classification \cite{mitchellmax}. In practice, cross-entropy can lead to suboptimal results due to several reasons. First, we usually don't have to predict true posteriors in order to achieve maximum accuracy. Any model with logits of the true class exceeding other logits leads to the optimal performance. On the other hand, overfitting and local optima prevent cross-entropy training from true posterior prediction. The limitations of cross-entropy gave a rise to other classification losses.


One of the widely used classification losses is hinge loss \cite{gentile1998linear}.
Unlike cross-entropy, hinge loss stops training when scores of ground truth classes exceed alternative scores with the required margin. 
In some problems, hinge loss shows results on par or better than cross-entropy \cite{EPALLE2021107375, jin2014traffic, OZYILDIRIM2021564, peng2018discriminative}.


Loss functions, such as cross-entropy and hinge loss, correlate with accuracy but do not optimize it directly \cite{grabocka2019learning}. For this reason, these methods are referred to as proxy or surrogate losses. In this work, we propose a different approach that directly optimizes the accuracy of a specially designed stochastic model.


\subsection{Surrogate Losses Beyond Accuracy}
Many machine learning methods, especially deep learning approaches, rely on gradient descent optimization \cite{mitchellgrad}, which is applicable only to differentiable loss functions. Surrogate losses provide differentiable approximations for non-differentiable target metrics. However, optimization of surrogate losses does not necessary leads to a target metric optimization. Previous works propose differentiable surrogate losses for metrics beyond accuracy, including ROC AUC \cite{calders2007efficient, yuan2021large} and F1 scores \cite{benedict2021sigmoidf1}. For example, AUC ROC can be rewritten in terms of the Heaviside step function, which is approximated via a logistic function or polynomials. Unlike the above-mentioned approaches, we focus on a direct target metric optimization.

Surrogate loss functions were proposed in domains such as metric learning and ranking \cite{kaya2019metricsurvey, calauzenes2012surrogaterank}. Many of them use margin-based losses, similar to hinge loss \cite{weinberger2009triplet}. Some surrogate losses formulate target metrics in terms of the Heaviside step function, which is then approximated by smooth functions \cite{patel2021recallsurrogate}.

The proposed EXACT method is designed for direct accuracy optimization and does not require any surrogate loss function. Furthermore, the idea of using stochastic prediction can potentially be extended beyond accuracy optimization and applied in other machine learning domains.


\subsection{Stochastic Prediction}

There are multiple ways to introduce stochasticity to the model's output. One way is to use probabilistic embeddings \cite{shi2019probabilistic}, where the model predicts a distribution of outputs rather than a single vector. For example, the model can predict a mean and covariance matrix of the multivariate normal distribution. This concept was used in many deep learning approaches, such as regression via mixture density networks \cite{bishop1994mixture}, variational auto-encoders \cite{kingma2014vae}, and metric learning \cite{chang2020data}. Another way to introduce stochasticity is to use Bayesian neural networks (BNNs) \cite{goan2020bayesian}. In BNNs, the model weights are treated as random variables, and the output depends on particular weights' values.


One part of the proposed EXACT method is stochastic prediction, which is similar to probabilistic embeddings. However, unlike previous approaches, we use stochastic prediction for accuracy optimization.




\section{Motivation}

Let's consider a toy example where cross-entropy and hinge loss minimization produces suboptimal results in terms of accuracy. Suppose we solve a binary classification problem via the simple threshold model $f(x) = x - b$. The model predicts class $\tilde{y}$ based on the sign of $f(x)$:
\begin{equation}
    \tilde{y} = \begin{cases}
      -1, & f(x) < 0 \\
      1, & f(x) \ge 0
    \end{cases}.
\end{equation}
The goal of training is to fit the threshold parameter $b$. Suppose the training set consists of three points: -0.25, 0, and 0.25. The first two points have the label $y = -1$, and the last point has the label $1$, as shown in Figure \ref{fig:toy}.

Cross-entropy loss for binary classification is usually applied to the output of the logistic function $g$, which converts $f(x)$ to the probability of the positive class:
\begin{align}
    \mathcal{L}_{CE}(x, y) &= -\log g(y f(x)), \\
    g(x) &= \frac{1}{1 + e^{-x}}.
\end{align}
Binary hinge loss is computed as follows:
\begin{equation}
    \mathcal{L}_{Hinge}(x, y) = \max(0, 1 - y f(x)).
\end{equation}

Minimization of cross-entropy and hinge losses fails to find the optimal value of $b$, which must be in the interval $(0, 0.25)$. The global minimum of the cross-entropy loss is $b = 0.7$, and $0.75 \le b \le 1$ for hinge loss. Gradient ascent optimization of the proposed EXACT method, which we describe below, achieves perfect classification in this example, leading to $b = 0.125$.

\section{EXACT}
The accuracy metric is not differentiable and cannot be directly optimized via gradient ascent. The core idea behind EXACT is to introduce stochasticity to the model's output and then optimize the expected accuracy, i.e. accuracy of the stochastic model.

\subsection{Definitions}
The goal of classification training is to minimize the empirical risk of the mapping $x \longrightarrow \tilde{y}$, where $x \in \mathbb{R}^d$ is an input feature vector and $\tilde{y} \in \overline{1, C}$ is an output class label. In machine learning with gradient methods, including deep learning, the problem is usually solved by a parametric differentiable function $f_\theta: \mathbb{R}^d \longrightarrow \mathbb{R}^C$ with parameters $\theta$, which predicts score vectors (or logits) of output classes. Prediction is obtained by taking the class with the maximum score:
\begin{equation}
    \tilde{y}_\theta(x) = \argmax\limits_{i \in \overline{1, C}}f_\theta(x)_i.
\end{equation}
In some equations below we will omit $\theta$ subscript for simplicity.

In this work, we consider accuracy maximization, i.e. maximization of the correct classification probability of elements $(x, y)$ from some dataset:
\begin{equation}
    \mathcal{A}(\theta) = \mathbb{E}_{x, y}\mathbbm{1}(\tilde{y}_\theta(x) = y).
\end{equation}
\begin{equation}
    \hat{\theta} = \argmax\limits_\theta \mathcal{A}(\theta),
\end{equation}
where $\hat{\theta}$ is an optimal set of parameters, which we estimate using gradient ascent.

\subsection{Stochastic Model's Accuracy}

\begin{figure}[t]
\vskip 0.1in
\centering
\includegraphics{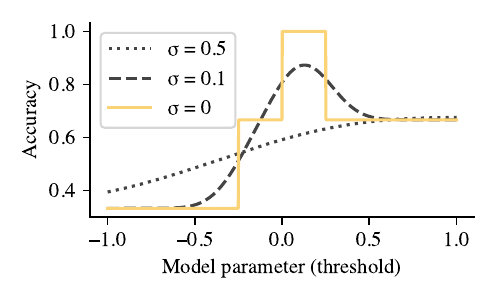}
\caption{Dependency of the expected accuracy on the model parameter in our toy example for different values of $\sigma$.}
\label{fig:sigma}
\vskip -0.1in
\end{figure}

Let's consider a modification of the function $f: x \rightarrow (\mu, \sigma)$ with $\sigma > 0$, which predicts the multivariate normal distribution of a score vector $s$ rather than a single vector:
\begin{equation}
    s \sim \mathcal{N}(\mu(x), \sigma^2(x) I),
    \label{eq:scores-distribution}
\end{equation}
\begin{equation}
    \tilde{y}_\theta(s) = \argmax\limits_{i \in \overline{1, C}} s_i.
    \label{eq:argmax}
\end{equation}
In this case, the predicted class $\tilde{y}_\theta(s)$ is a discrete random variable, which depends on the value of the score vector $s$. We call such a model stochastic since it produces random output. Accuracy of the stochastic model is an expectation of accuracy w.r.t. values of the score vector:
\begin{equation}
    \mathcal{A}(\theta) = \mathbb{E}_{x, y}\mathbb{E}_{s}\mathbbm{1}(\tilde{y}_\theta(s) = y).
\end{equation}
Since the indicator function is a Bernoulli random variable, we can replace the indicator's expectation with probability:
\begin{equation}
    \mathcal{A}(\theta) =  \mathbb{E}_{x, y}\mathrm{P}(\tilde{y}_\theta(s) = y).
\end{equation}
By taking into account Equation \ref{eq:argmax}, we can rewrite accuracy as
\begin{equation}
    \mathcal{A}(\theta)
    = \mathbb{E}_{x, y}\mathrm{P}(s_y > \max\limits_{i \ne y}s_i).
    \label{eq:total-accuracy}
\end{equation}
Here we assume that classification is incorrect if scores of two or more classes are equal to each other.

Examples of the expected accuracy for different values of $\sigma$ are presented in Figure \ref{fig:sigma}. As we will show below, the accuracy of the stochastic model is differentiable w.r.t. mean vector $\mu$. This property makes optimization via gradient ascent possible.

\subsection{Optimization}
In this section, we propose an effective approach to expected accuracy computation and differentiation. We base this method on reducing the original problem to an orthant integration of the multivariate normal probability density function (PDF). The latter can be effectively computed using algorithms proposed in previous works.

\vskip 0.05in
\begin{definition}
Delta matrix $D_y$ of the order $C$ for the ground truth label $y$ is a matrix of size $C - 1 \times C$ with
\begin{equation}
    {D_y}_{i, j} = \begin{cases}
      1, & j = y \\
      -1, & j < y, i = j \\
      -1, & j > y, i = j - 1 \\
      0, & otherwise
    \end{cases}.
\end{equation}
\end{definition}
In other words, the delta matrix is used to compute differences between scores of different classes and has the following form:
\begin{equation}
D_y = \begin{bmatrix}
  -1 & 0 & \dots & 0 & 1 & 0 & \dots & 0 \\
  0 & -1 & \dots & 0 & 1 & 0 & \dots & 0 \\
  & & \ddots & & \vdots & & \ddots & \\
  0 & 0 & \dots & -1 & 1 & 0 & \dots & 0 \\
  0 & 0 & \dots & 0 & 1 & -1 & \dots & 0 \\
  & & \ddots & & \vdots & & \ddots & \\
  0 & 0 & \dots & 0 & 1 & 0 & \dots & -1
\end{bmatrix}.
\end{equation}

Now we can state a key result, necessary for expected accuracy evaluation.
\vskip 0.05in
\begin{theorem}
Suppose the scores vector $s$ is distributed according to multivariate normal distribution $\mathcal{N}(\mu, \sigma^2 I)$ in $\mathbb{R}^C$. In this case, the probability of the $y$-th score exceeding other scores can be represented as
\begin{equation}
    \mathrm{P}(s_y > \max\limits_{i \ne y} s_i) = \int\limits_{\Omega_+}\mathcal{N}(t; \frac{D_y\mu}{\sigma}, D_yD_y^T) dt,
    \label{eq:exact-comp}
\end{equation}
where $\mathcal{N}(t; \mu, \Sigma)$ denotes multivariate normal PDF, $D_y$ is a delta matrix of the order $C$ for the label $y$ and $\Omega_+: \{t \in \mathbb{R}^{C-1}, t_i \ge 0, i = \overline{1,C-1}\}$ is an orthant in $\mathbb{R}^{C-1}$.
\end{theorem}
The orthant integral in equation \ref{eq:exact-comp} can be effectively computed using Genz algorithm \cite{genz1992numerical}. The Equation \ref{eq:exact-comp} represents correct classification probability for a single element of the dataset. Total accuracy is computed according to equation \ref{eq:total-accuracy} as an average of correct classification probabilities for all elements.


Gradient ascent requires gradient computation w.r.t. parameters $\mu$ and $\sigma$. Both parameters are included only in the mean part of the normal distribution in Equation \ref{eq:exact-comp}. Let's define
\begin{align}
    \label{eq:mean-part}
    m &= \frac{D_y\mu}{\sigma}, \\
    \Sigma &= D_yD_y^T.
\end{align}
Note, that the matrix $\Sigma$ is constant for each element of the dataset and doesn't require gradients. Note also that $\Sigma$ is a covariance matrix by construction and thus can be used as a parameter of a multivariate normal distribution. Gradients w.r.t. mean vector $m$ can be computed using the following theorem.
\vskip 0.1in
\begin{theorem}
\label{theor:grad}
Suppose the scores vector $s$ is distributed according to multivariate normal distribution $\mathcal{N}(m, \Sigma)$ in $\mathbb{R}^{C-1}$ and $\hat s_i$ is a random vector of all elements in $s$ except $i$-th, conditioned on $s_i = -m_i$. Then $\hat s_i$ is normally-distributed with some parameters $\hat m_i$ and $\hat \Sigma_i$ and
\begin{multline}
    \left(\int\limits_{\Omega_+}\mathcal{N}(t; m, \Sigma) dt\right)'_{m_i} \\  = \mathcal{N}(0;m_i, \Sigma_{i, i})
    \int\limits_{\Omega'_+}\mathcal{N}(t; \hat m_i, \hat \Sigma_i) dt,
    \label{eq:exact-grad}
\end{multline}
where $\mathcal{N}(t; \mu, \Sigma)$ denotes multivariate normal PDF and $\Omega'_+: \{t \in \mathbb{R}^{C-2}, t_i \ge 0, i = \overline{1,C-2}\}$ is an orthant in $\mathbb{R}^{C-2}$.
\end{theorem}
Same as before, the orthant integral is computed using Genz algorithm \cite{genz1992numerical}.
According to Theorem \ref{theor:grad} and Equation \ref{eq:mean-part}, accuracy of the stochastic model is differentiable w.r.t. both $\mu(x)$ and $\sigma(x)$. The gradient descent method can be applied to minimize the negative value of accuracy.



\begin{figure}[t]
\vskip 0.1in
\centering
\includegraphics[width=.9\columnwidth]{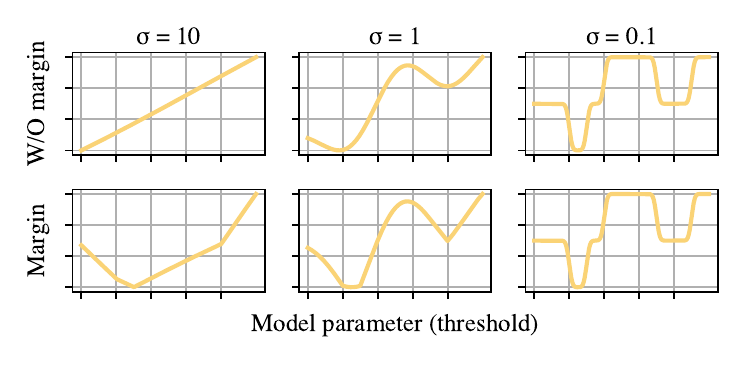}
\caption{EXACT loss dependency on the model parameter with and w/o margin. Margin affects training with large $\sigma$, creating a better optimization landscape in early epochs.}
\label{fig:margin}
\vskip -0.in
\end{figure}

\subsection{Inference}

During inference, EXACT predicts a score vector with maximum density, i.e. $\mu$ vector. Variance prediction can be excluded from the model in inference time, leading to zero computational overhead.

\subsection{Improvements}
In this section, we analyze corner cases and training with \mbox{EXACT} loss and propose improvements aimed at training stabilization and better final optima.

\subsubsection{Margin}
When $\sigma$ is close to zero, the stochastic model's accuracy reduces to a single-point prediction. In this case \mbox{EXACT} loss resembles the accuracy of the traditional model and suffers from local accuracy optima. To achieve better optimum during training, EXACT has to start with large $\sigma$. However, if $\sigma$ is very large, then the surface of EXACT loss function will interpolate prior class probabilities, as shown in Figure \ref{fig:margin}. In the latter case, EXACT completely smooths optima. To make the method more robust to large initial $\sigma$ values, we incorporate margin to the loss function.
EXACT margin is similar to that of hinge loss and affects the computation of the mean vector from Equation \ref{eq:mean-part}:
\begin{equation}
    m_r = \frac{\min\left(D_y\mu, r\right)}{\sigma},
\end{equation}
where $\min$ is applied element-wise and $r$ is a margin value. The obtained $m_r$ value is then used instead of $m$ in Equation \ref{eq:mean-part} for loss and gradient computation. As shown in Figure \ref{fig:margin}, margin affects training with large $\sigma$ values, while has almost no effect as $\sigma$ decreases.
\begin{table*}[t]
\centering
\begin{tabular}{lc|cccc}
Dataset & Split & Sklearn & Cross-entropy & Hinge & EXACT \\
\hline

adult                   & train & $85.33 \pm 0.00$ & $85.38 \pm 0.02$ & $85.28 \pm 0.01$ & $\bf 85.47 \pm 0.01$ \\
annealing               & train & $\bf 100.00 \pm 0.00$ & $\bf 100.00 \pm 0.00$ & $\bf 100.00 \pm 0.00$ & $\bf 100.00 \pm 0.00$ \\
audit-risk              & train & $99.19 \pm 0.00$ & $98.23 \pm 0.00$ & $\bf 100.00 \pm 0.00$ & $\bf 100.00 \pm 0.00$ \\
balance-scale           & train & $91.40 \pm 0.00$ & $91.40 \pm 0.00$ & $\bf 91.60 \pm 0.00$ & $\bf 91.60 \pm 0.00$ \\
breast-cancer-wisconsin & train & $98.68 \pm 0.00$ & $98.02 \pm 0.00$ & $98.90 \pm 0.00$ & $\bf 99.03 \pm 0.11$ \\
car                     & train & $94.50 \pm 0.00$ & $94.50 \pm 0.00$ & $93.63 \pm 0.00$ & $\bf 96.67 \pm 0.00$ \\
cylinder-bands          & train & $80.05 \pm 0.00$ & $80.28 \pm 0.00$ & $79.86 \pm 0.09$ & $\bf 87.42 \pm 0.17$ \\
dry-bean                & train & $\bf 92.79 \pm 0.00$ & $92.38 \pm 0.00$ & $92.44 \pm 0.01$ & $92.67 \pm 0.02$ \\
mice-protein            & train & $\bf 100.00 \pm 0.00$ & $\bf 100.00 \pm 0.00$ & $99.44 \pm 0.11$ & $\bf 100.00 \pm 0.00$ \\
wine                    & train & $\bf 100.00 \pm 0.00$ & $\bf 100.00 \pm 0.00$ & $\bf 100.00 \pm 0.00$ & $\bf 100.00 \pm 0.00$ \\
\hline
Total best & & 4 & 3 & 4 & 9 \\

\end{tabular}
\vskip 0.2in

\caption{Train set accuracy (\%) of linear models trained with different loss functions on 10 tabular datasets. Mean and STD of 5 runs with different seeds are reported. Sklearn training doesn't depend on the random seed and thus STD is always zero.}
\label{tab:uci-train}
\end{table*}

\begin{table*}[t]
\centering
\begin{tabular}{lc|cccc}
Dataset & Split & Sklearn & Cross-entropy & Hinge & EXACT \\
\hline

adult                   & test & $\bf 85.28 \pm 0.00$ & $\bf 85.28 \pm 0.01$ & $85.26 \pm 0.01$ & $85.21 \pm 0.01$ \\
annealing               & test & $\bf 100.00 \pm 0.00$ & $\bf 100.00 \pm 0.00$ & $\bf 100.00 \pm 0.00$ & $\bf 100.00 \pm 0.00$ \\
audit-risk              & test & $\bf 98.72 \pm 0.00$ & $96.28 \pm 0.48$ & $97.05 \pm 0.65$ & $96.67 \pm 0.75$ \\
balance-scale           & test & $\bf 92.00 \pm 0.00$ & $\bf 92.00 \pm 0.00$ & $\bf 92.00 \pm 0.00$ & $\bf 92.00 \pm 0.00$ \\
breast-cancer-wisconsin & test & $95.61 \pm 0.00$ & $95.96 \pm 0.43$ & $96.49 \pm 0.00$ & $\bf 97.37 \pm 0.00$ \\
car                     & test & $92.20 \pm 0.00$ & $91.91 \pm 0.00$ & $92.77 \pm 0.00$ & $\bf 95.20 \pm 0.14$ \\
cylinder-bands          & test & $72.22 \pm 0.00$ & $72.78 \pm 0.45$ & $74.44 \pm 0.45$ & $\bf 77.22 \pm 0.45$ \\
dry-bean                & test & $\bf 93.06 \pm 0.00$ & $92.83 \pm 0.01$ & $92.93 \pm 0.03$ & $92.99 \pm 0.07$ \\
mice-protein            & test & $\bf 99.07 \pm 0.00$ & $98.89 \pm 0.23$ & $97.69 \pm 0.00$ & $97.69 \pm 0.00$ \\
wine                    & test & $\bf 100.00 \pm 0.00$ & $\bf 100.00 \pm 0.00$ & $\bf 100.00 \pm 0.00$ & $\bf 100.00 \pm 0.00$ \\
\hline
Total best & & 7 & 4 & 3 & 6 \\

\end{tabular}
\vskip 0.2in
\caption{Test set accuracy (\%) of linear models trained with different loss functions on 10 tabular datasets. Mean and STD of 5 runs with different seeds are reported. Sklearn training doesn't depend on the random seed and thus STD is always zero.}
\label{tab:uci-test}
\end{table*}

\begin{table*}[t]
\centering
\vskip 0.2in
\begin{tabular}{l|l|cccc}
\multirow{2}{*}{Method} & \multirow{2}{*}{Optimizer} & MNIST & SVHN & CIFAR-10 & CIFAR-100 \\
& & \scriptsize M3-CNN & \scriptsize Wide ResNet 16-8 & \scriptsize Wide ResNet 28-10 & \scriptsize Wide ResNet 28-10 \\
\hline
Cross-entropy & \multirow{3}{*}{SGD} & 99.64 ± 0.03 & \bf 97.56 ± 0.03 & 95.54 ± 0.09 & 80.18 ± 0.2 \\
Hinge & & \bf 99.68 ± 0.04 & 97.06 ± 0.04 & 94.36 ± 0.09 & 79.42 ± 0.4 \\
EXACT & & 99.67 ± 0.05 & 97.47 ± 0.06 & \bf 96.33 ± 0.13 & \bf 80.71 ± 0.17 \\
\hline
Cross-entropy & \multirow{3}{*}{Adam}
        & 99.58 ± 0.04 & 96.14 ± 0.17 & \bf 94.26 ± 0.16 & 71.31 ± 0.37 \\
Hinge & & \bf 99.61 ± 0.04 & 95.85 ± 0.22 & 93.17 ± 0.14 & 58.03 ± 1.14 \\
EXACT & & 99.58 ± 0.08 & \bf 96.76 ± 0.10 & 93.16 ± 0.18 & \bf 72.00 ± 0.31 \\
\hline
Cross-entropy & \multirow{3}{*}{ASAM} & 99.65 ± 0.04 & 97.46 ± 0.05 & 96.23 ± 0.18 & 82.40 ± 0.19 \\
Hinge & & \bf 99.71 ± 0.02 & 96.34 ± 0.04 & 95.45 ± 0.17 & 77.66 ± 0.30\\
EXACT & & 99.67 ± 0.03 & \bf 97.79 ± 0.02 & \bf 96.73 ± 0.08 & \bf 82.68 ± 0.10 \\
\end{tabular}
\vskip 0.2in
\caption{Test set accuracy (\%) in deep vision tasks for top-performing neural architectures with different optimizers. Mean and STD of 5 runs with different seeds are reported.}
\label{tab:vision}
\end{table*}

\subsubsection{Ratio Ambiguity}
According to Equation \ref{eq:exact-comp}, $\mu$ and $\sigma$ are used in EXACT loss only as a fraction $\frac{\mu}{\sigma}$. This leads to an ambiguity in the model's prediction because the multiplication of both values by the same number will not affect stochastic accuracy. To decouple the effects of $\mu$ and $\sigma$, we apply batch normalization \cite{ioffe2015batch} to the $\mu$ branch of the model. Batch normalization is a part of the loss function and is removed during inference.

\begin{figure}[t]
\vskip 0.0in
\centering
\includegraphics[width=.9\columnwidth]{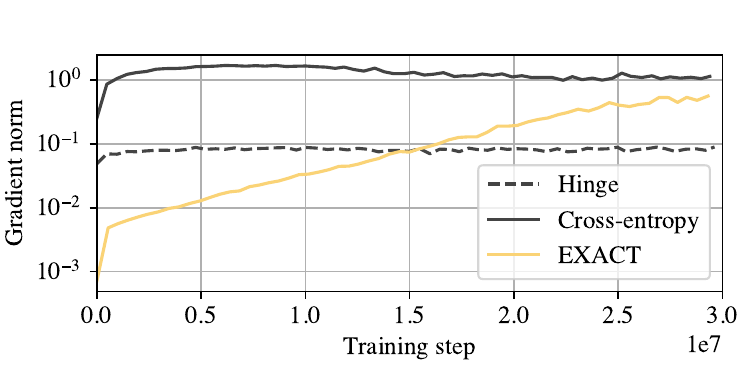}
\caption{Gradient norm during training on CIFAR-100 for different loss functions.}
\label{fig:gradnorm}
\vskip -0.2in
\end{figure}

\subsubsection{Variance Scheduler}
While $\sigma$ prediction subnetwork can be trained simultaneously with other parameters, we found it more beneficial to change $\sigma$ according to the predefined schedule. One reason behind this is that gradient descent rapidly reduces predicted variance, converging to the closest (usually poor) local optimum. Slow reduction of the variance forces the algorithm to make more steps with large $\sigma$ and results in higher final accuracy.

\subsubsection{Gradient Normalizer}
In practice, EXACT gradient norm largely depends on the output variance. As shown in Figure \ref{fig:gradnorm}, cross-entropy and hinge losses achieve plateau during training, while EXACT gradient norm exponentially growth. To better control the update step, we divide EXACT's gradient by running mean statistics. The running mean is computed using exponential smoothing with a factor 0.9.

\section{Experimental Setup}
In this section, we describe data preparation and training details of the proposed method and the baselines.

\subsection{Model Architectures}
For tabular data classification, we use linear models. Scores (or logits) of the first class are fixed to zero. Scores of the remaining classes are predicted via linear transform \mbox{$Ax + b$} of input features $x$. This setup is equivalent to a binary classification model when the number of classes is equal to 2. We also evaluate logistic regression implementation from Sklearn \cite{scikit-learn}.

For image classification datasets we use different neural network models to better match data complexity. For MNIST we use 10-layer plain M3 CNN architecture from the recent work \cite{an2020simplecnn}, which achieved top-performing results on this dataset. For SVHN and CIFAR we use Wide ResNet (WRN) architecture \cite{zagoruyko2016wide}, which achieved better performance than ResNet \cite{he2016deep}. Particularly, we use WRN-16-8 for SVHN and WRN-28-10 for CIFAR-10 / CIFAR-100.

\subsection{Tabular Data Preparation}
We evaluate linear models using datasets from UCI machine learning repository \cite{Dua:2019}. We choose 10 popular classification datasets with numeric and categorical features. Most UCI datasets have missing values. In order to apply linear models to these datasets, we fill missing numeric data with mean values and missing categorical data with most frequent elements. We also apply one-hot encoding to categorical data. All features are normalized using per-feature mean and STD statistics. For datasets without original test part, we split 20\% randomly selected items for testing.

\subsection{Image Data Preparation}
We conduct experiments on MNIST \cite{lecun1998mnist}, SVHN \cite{netzer2011svhn}, CIFAR-10, and CIFAR-100 \cite{krizhevsky2009cifar} datasets. We prefer small and medium-size datasets because hyperparameter tuning on large datasets is practically unattainable.

For MNIST we use simple data transform consisting of random translation of the image by at most 20\% of the original size and random rotation by at most 20 degrees \cite{an2020simplecnn}. For SVHN we use auto augmentation \cite{cubuk2019autoaugment} from PyTorch \cite{NEURIPS2019_9015} with parameters designed for CIFAR-10. For CIFAR-10 and CIFAR-100 we use CIFAR-10 auto augmentations along with random horizontal flipping.

All input images are scaled to the fixed size depending on the neural architecture to achieve spatial tensor size \mbox{$8 \times 8$} before the final pooling layer. For M3, used for MNIST, images are scaled to 28 pixels on each side. For Wide ResNet architectures, used for SVHN, CIFAR-10, and CIFAR-100, images are scaled to 32 pixels on each side.

\subsection{Hyperparameters and Training}

The list of hyperparameters includes initial learning rate, gradient clipping threshold, and margin (for hinge loss and EXACT). Hyperparameters are tuned via 50 runs of random search with 20\% of the training set reserved for validation. For linear models we also tune L2 regularization coefficient.

All models, except Sklearn, are trained with batch size equal to 256. Linear models are trained for 8000 steps to balance training for different dataset sizes. The logistic regression from Sklearn is trained using standard L-BFGS solver \cite{byrd1995limited}. Neural networks are trained for 150 epochs on MNIST and SVHN and for 500 epochs on CIFAR10 and CIFAR100. Models are trained via stochastic gradient descent with momentum equal to $0.9$ and weight decay equal to $10^{-4}$. Learning rate exponentially decays from initial value to $10^{-4}$ at last epoch.  We also train neural networks with Adam \cite{kingma2015adam} and ASAM \cite{kwon2021asam} optimizers. EXACT variance scheduler reduces $\sigma$ from 10 in the first epoch to the final value, which is 0.01 for linear models and 1 for neural networks.

We use the PyTorch deep learning framework \cite{paszke2019pytorch} for all models except logistic regression from Sklearn. Each experiment is performed using a single NVIDIA V100 GPU.

EXACT implementation depends on the sample size used in the Genz integration algorithm. We use sample size 16 in all comparisons.

\section{Experiments}
In this work, we aim to answer the following questions: (1) how does EXACT performance correspond to that of cross-entropy and hinge losses, (2) how to choose EXACT sample size, and (3) how big is the computational overhead of the proposed EXACT method? The answers to these questions are summarized in the subsections below.

\subsection{Classification Quality}

\subsubsection{Linear Models}
We compared linear models trained with Sklearn, cross-entropy loss, hinge loss, and the proposed EXACT loss. The comparison on 10 tabular datasets is presented in Tables \ref{tab:uci-train} and \ref{tab:uci-test}. On 9 out of 10 datasets EXACT achieves the highest accuracy on the training part of the dataset. It demonstrates the effectiveness of the proposed loss function in accuracy maximization. During testing \mbox{EXACT} achieves the highest accuracy in 6 out of 10 cases. On 3 benchmarks EXACT strictly outperforms all other methods, including Sklearn's logistic regression trained with L-BFGS solver. 

\subsubsection{Deep Image Classification}
We compared EXACT with surrogate losses on multiple datasets with different optimizers. The results are presented in Table \ref{tab:vision}. Hinge loss achieves the highest accuracy on MNIST dataset. On SVHN and CIFAR-10, results are mixed, but EXACT outperforms other methods in most cases. On CIFAR-100 EXACT strictly outperforms all baselines. Generally, EXACT outperforms other methods in 7 out of 12 comparisons.



\subsection{Sample Size}
\begin{figure}[t]
\vskip 0.2in
\centering
\includegraphics[width=\columnwidth]{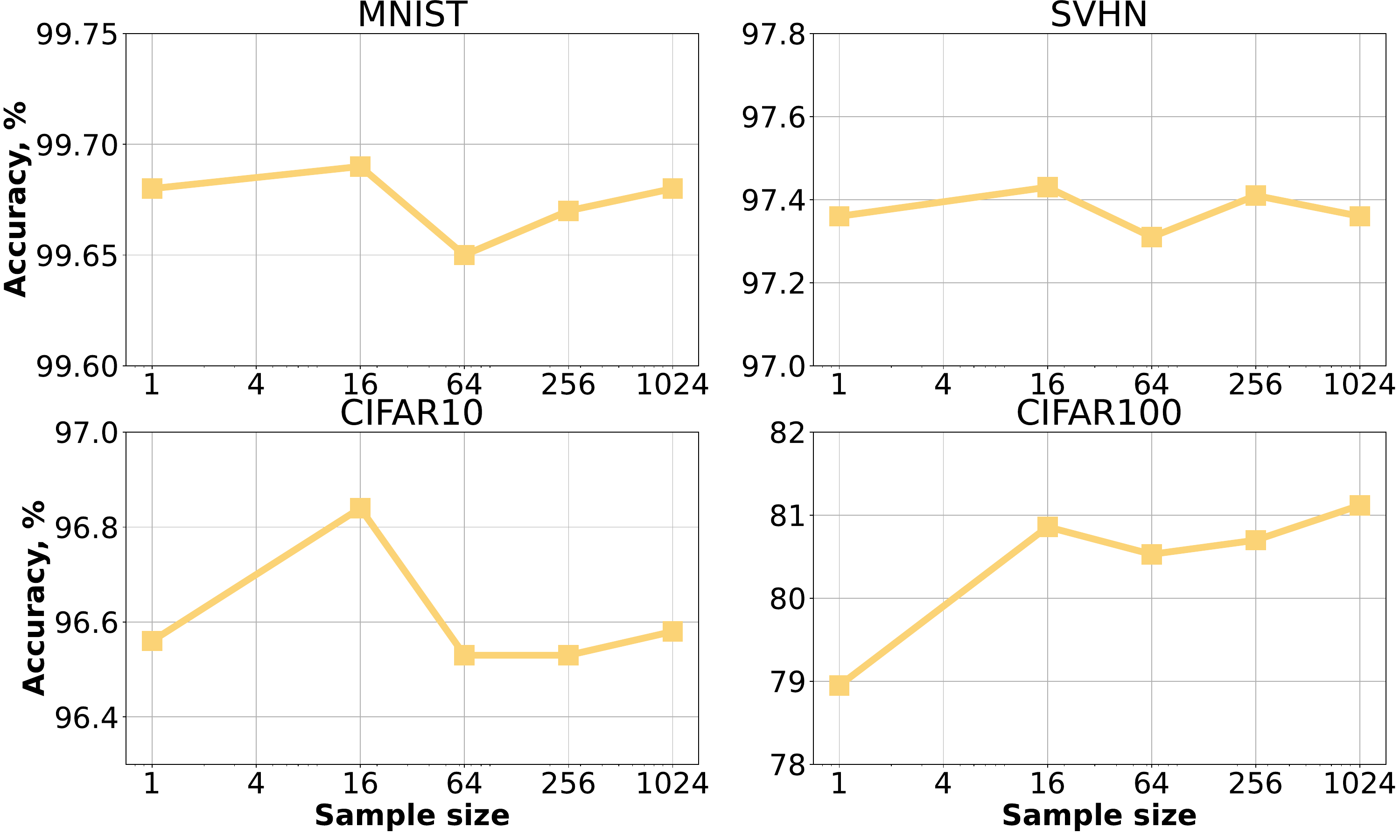}
\caption{EXACT test accuracy dependency on the sample size on different datasets. Results are presented for SGD optimizer.}
\label{fig:samplesize}
\vskip -0.2in
\end{figure}

EXACT depends on the sample size used in the Genz integration algorithm \cite{genz1992numerical}. Training results for different sample sizes are presented in Figure \ref{fig:samplesize}. On MNIST, SVHN, and CIFAR-10, the sample size has minor effects on accuracy. Even training with sample size 1 produces results on par with larger samples. On CIFAR-100, a larger sample size generally leads to higher accuracy.

\subsection{Computational Complexity}
\begin{figure}[t]
\vskip 0.2in
\centering
\includegraphics[width=\columnwidth]{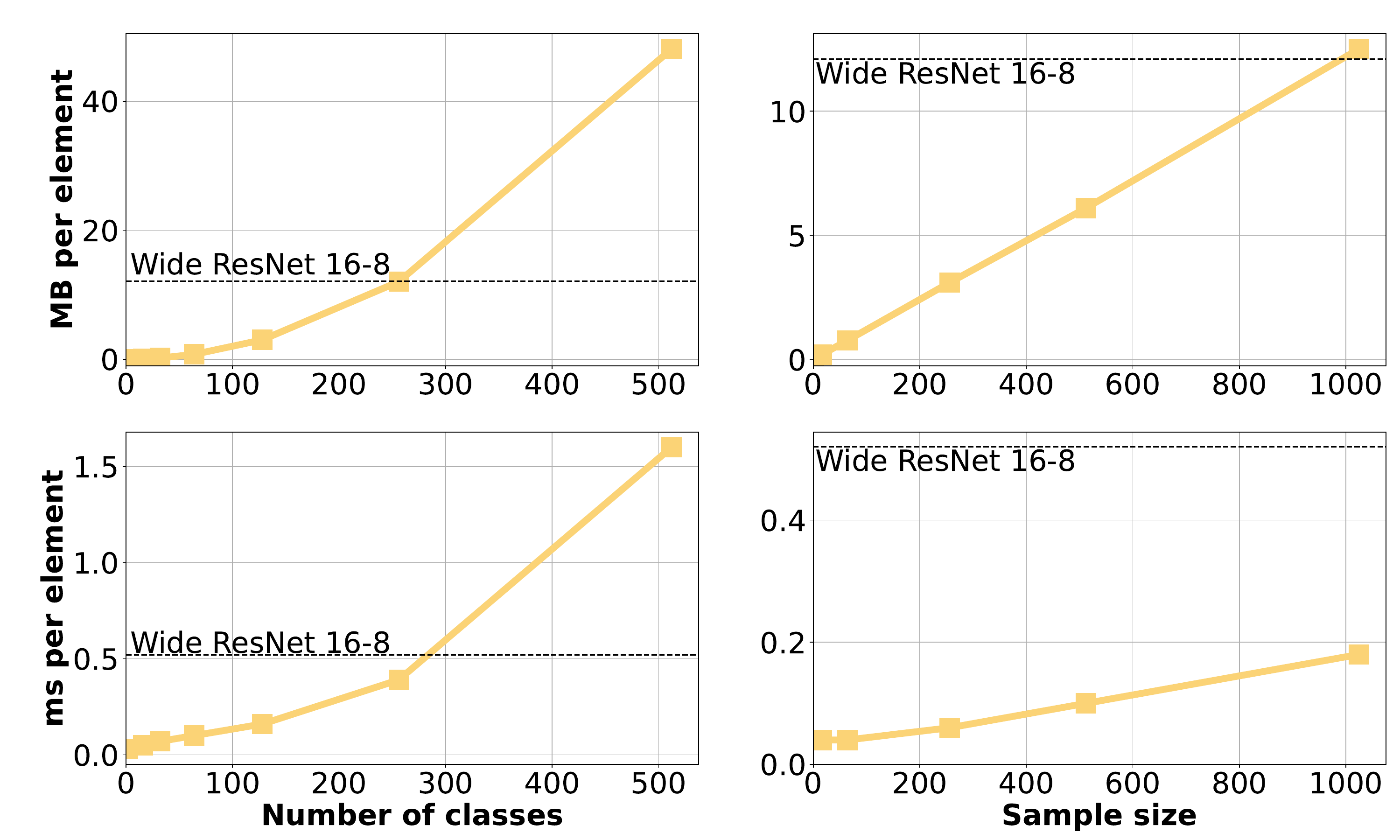}
\caption{EXACT loss memory consumption (MB per element) and computation speed (ms per element) for different numbers of classes and sample sizes. Performance is compared to the Wide ResNet 16-8 neural network.}
\label{fig:timememory}
\vskip -0.2in
\end{figure}

EXACT memory consumption and computational efficiency depend on the number of classes $C$ and sample size $N$. The most expensive part of the algorithm is gradient computation, which requires $C$ calls to Genz algorithm with complexity $O(N C)$. Thus the total complexity of EXACT loss is $O(N C^2)$ in terms of both operations and memory. Empirical memory usage per data element and computational efficiency (milliseconds per data element) are presented in Figure \ref{fig:timememory}. Evaluations for different numbers of classes were made with a sample size equal to 16. Different sample sizes were evaluated with the number of classes equal to 32.

The relative EXACT overhead largely depends on the number of classes and model architecture. For linear models EXACT overhead can achieve $200\%$. For neural networks overhead is much lower. For example, EXACT with the sample size equal to 16 in the case of 10 classes reduces training RPS by $3.7\%$ for Wide ResNet 16-8. For larger models, relative loss overhead is smaller and vice versa. For example, overhead for Wide ResNet 28-10 in the same setup is $0.3\%$. Wide ResNet 28-10 with 100 classes results in overhead of $5.5\%$.
\section{Discussion}
While previous works on deep classification minimized surrogate losses, EXACT directly optimizes the accuracy of the model with stochastic prediction.
The benefits of accuracy optimization were illustrated in our toy example, where both cross-entropy and hinge losses failed to solve the problem.
The proposed EXACT method treats non-differentiable accuracy metric in a novel way which can potentially be applied to domains beyond classification, such as metric learning, ranking, etc.

According to our experiments, EXACT leads to competitive results in tabular and image classification tasks. In many cases, EXACT achieves higher classification accuracy with a computational overhead of about $0-6\%$ for neural networks. While computational efficiency is dependent on sample size, we show that EXACT can be trained even with single-point estimation. Extra computational resources can be used to further increase EXACT accuracy.
\section{Conclusion}
We presented EXACT, a novel approach for optimizing stochastic model accuracy via gradient ascent. Our results show that EXACT achieves higher accuracy than cross-entropy and hinge losses in several tabular and image classification tasks, including SVHN, CIFAR10, and CIFAR100. EXACT can be effectively implemented in popular deep learning frameworks, including PyTorch, and leads to the computational overhead of about $3\%$ depending on the number of classes and neural model complexity.

\bibliography{exact}
\bibliographystyle{style}

\newpage
~
\newpage

\theoremstyle{plain}
\newtheorem{manualtheoreminner}{Theorem}
\newenvironment{manualtheorem}[1]{%
  \renewcommand\themanualtheoreminner{#1}%
  \manualtheoreminner
}{\endmanualtheoreminner}

\appendix
\section{Proofs of Theorems}
\begin{manualtheorem}{4.2}
Suppose the scores vector $s$ is distributed according to multivariate normal distribution $\mathcal{N}(\mu, \sigma^2 I)$ in $\mathbb{R}^C$. In this case, the probability of the $y$-th score exceeding other scores can be represented as
\begin{equation}
    \mathrm{P}(s_y > \max\limits_{i \ne y} s_i) = \int\limits_{\Omega_+}\mathcal{N}(t; \frac{D_y\mu}{\sigma}, D_yD_y^T) dt,
    \label{eq:app-exact-comp}
\end{equation}
where $\mathcal{N}(t; \mu, \Sigma)$ denotes multivariate normal PDF, $D_y$ is a delta matrix of the order $C$ for the label $y$ and $\Omega_+: \{t \in \mathbb{R}^{C-1}, t_i \ge 0, i = \overline{1,C-1}\}$ is an orthant in $\mathbb{R}^{C-1}$.
\end{manualtheorem}
\begin{proof}
Let's rewrite Equation \ref{eq:app-exact-comp}:
\begin{align}
    \mathrm{P}(s_y > \max\limits_{i \ne y} s_i)
    &= \mathrm{P}(s_y - \max\limits_{i \ne y} s_i > 0) \\
    &= \mathrm{P}(\min\limits_{i \ne y}( s_y - s_i) > 0).
\end{align}
Due to the definition of the delta matrix $D_y$, vector $d = D_ys$ is a $(C - 1)$-dimensional vector with elements equal to differences between the score of the ground truth class $y$ and other scores:
\begin{equation}
    d_i = \begin{cases}
    s_y - s_i, & i < y \\
    s_y - s_{i + 1}, & i >= y
    \end{cases},
\end{equation}
then
\begin{align}
    \mathrm{P}(s_y > \max\limits_{i \ne y} s_i)
    &= \mathrm{P}(\min\limits_i d_i > 0) \\
    &= \mathrm{P}(\min\limits_i \frac{d_i}{\sigma} > 0) \\
    &= \mathrm{P}(\min\limits_i w_i > 0),
\end{align}
where $w = \frac{D_y s}{\sigma}$. Due to the properties of the multivariate normal distribution, random vector $w$ is also normally distributed:
\begin{equation}
    w \sim \mathcal{N}(\frac{D_y\mu}{\sigma}, D_yD_y^T).
\end{equation}
Finally:
\begin{align}
    \mathrm{P}(s_y > \max\limits_{i \ne y} s_i)
    &= \mathrm{P}(w_1 > 0, \dots, w_{C - 1} > 0) \\
    &= \int\limits_{\Omega_+}\mathcal{N}(t; \frac{D_y\mu}{\sigma}, D_yD_y^T) dt.
\end{align}
\end{proof}

\newpage
\begin{manualtheorem}{4.3}
\label{theor:app-grad}
Suppose the scores vector $s$ is distributed according to multivariate normal distribution $\mathcal{N}(m, \Sigma)$ in $\mathbb{R}^{C-1}$ and $\hat s_i$ is a random vector of all elements in $s$ except $i$-th, conditioned on $s_i = -m_i$. Then $\hat s_i$ is normally-distributed with some parameters $\hat m_i$ and $\hat \Sigma_i$ and
\begin{multline}
    \left(\int\limits_{\Omega_+}\mathcal{N}(t; m, \Sigma) dt\right)'_{m_i} \\  = \mathcal{N}(0;m_i, \Sigma_{i, i})
    \int\limits_{\Omega'_+}\mathcal{N}(t; \hat m_i, \hat \Sigma_i) dt,
    \label{eq:app-exact-grad}
\end{multline}
where $\mathcal{N}(t; \mu, \Sigma)$ denotes multivariate normal PDF and $\Omega'_+: \{t \in \mathbb{R}^{C-2}, t_i \ge 0, i = \overline{1,C-2}\}$ is an orthant in $\mathbb{R}^{C-2}$.
\end{manualtheorem}
\begin{proof}
\begin{align}
    \int\limits_{\Omega_+}\mathcal{N}&(t; m, \Sigma) dt \\
    &= \int\limits_0^\infty \dots \int\limits_0^\infty \mathcal{N}(t; m, \Sigma) dt_{C-1} \dots dt_1 \\
    &= \int\limits_0^\infty \dots \int\limits_0^\infty \mathcal{N}(t - m; 0, \Sigma) dt_{C-1} \dots dt_1 \\
    &= \int\limits_{-m_1}^\infty \dots \int\limits_{-m_{C-1}}^\infty \mathcal{N}(t; 0, \Sigma) dt_{C-1} \dots dt_1
\end{align}
Assume, without loss of generality, that $i = 1$. Then we can find the derivative using Leibniz integral rule:
\begin{align}
    &\left(\int\limits_{\Omega_+}\mathcal{N}(t; m, \Sigma) dt\right)'_{m_1} \\
    &= \left.\int\limits_{-m_2}^\infty \dots \int\limits_{-m_{C-1}}^\infty \mathcal{N}(t; 0, \Sigma) dt_{C-1} \dots dt_2 \right\vert_{t_1 = -m_1} \\
    &= \left.\int\limits_0^\infty \dots \int\limits_0^\infty \mathcal{N}(t; m, \Sigma) dt_{C-1} \dots dt_2 \right\vert_{t_1 = 0}
\end{align}
Integration region in the last integral is a positive orthant $\Omega'_+$, which is a lower dimension subset of $\Omega_+$:
\begin{multline}
    \left(\int\limits_{\Omega_+}\mathcal{N}(t; m, \Sigma) dt\right)'_{m_1} \\
    = \left.\int\limits_{\Omega'_+} \mathcal{N}(t; m, \Sigma) dt_{C-1} \dots dt_2 \right\vert_{t_1 = 0}.
    \label{eq:deriv}
\end{multline}
We can decompose inner density using properties of the multivariate normal distribution:
\begin{align}
    \mathrm{P}_s(t) &= \mathrm{P}_s(t_1, \dots, t_{C-1}) \\
    &= \mathrm{P}_s(t_1, t_{2:C-1}) \\
    &= \mathrm{P}_{s_1}(t_1) \mathrm{P}_{s_{2:C-1}}(t_{2:C-1} | s_1 = t_1),
    \label{eq:cond}
\end{align}
so
\begin{equation}
    \mathcal{N}(t; m, \Sigma) = \mathcal{N}(t_1;m_1, \Sigma_{1, 1})
    \mathcal{N}(t_{2:C-1}; \hat m_1, \hat \Sigma_1),
    \label{eq:decomp}
\end{equation}
where $\hat m_1$ and $\hat \Sigma_1$ are parameters of the conditional distribution from Equation \ref{eq:cond}:
\begin{align}
    \hat m_1 &= m_{2:C-1} + \Sigma_{2:C-1,1}\Sigma^{-1}_{1,1}(t_1 - m_1), \\
    \hat \Sigma_1 &= \Sigma_{2:C-1,2:C-1} - \Sigma_{2:C-1,1}\Sigma^{-1}_{1,1}\Sigma_{1,2:C-1}.
\end{align}
Substitution of the Equation \ref{eq:decomp} to Equation \ref{eq:deriv} finally proofs the theorem.
\end{proof}

\section{Implementation Notes}
EXACT loss computation algorithm with all improvements is presented in Listing \ref{lst:algorithm}.

\begin{listing}[h]%
\caption{Example PyTorch Code for EXACT computation}%
\label{lst:algorithm}%
\begin{lstlisting}
mu = empty(batch_size, dim)  # Input.
sigma = empty(batch_size)  # Input.
mu_bn = (mu - mu.mean()) / mu.std()
deltas = mu_bn @ Dy.T
deltas_margin = deltas.clip(max=r)
ratio = deltas_margin / sigma[:, None]
loss = 1 - genz_integral(mean=ratio,  cov=eye(dim) + 1)
\end{lstlisting}
\end{listing}

A considerable boost in EXACT performance can be obtained from analysis of the matrix $\Sigma = D_yD_y^T$ in Equation \ref{eq:app-exact-comp}. Matrix $D_y$ is a delta matrix for the label $y$:
\begin{equation}
D_y = \begin{bmatrix}
  -1 & 0 & \dots & 0 & 1 & 0 & \dots & 0 \\
  0 & -1 & \dots & 0 & 1 & 0 & \dots & 0 \\
  & & \ddots & & \vdots & & \ddots & \\
  0 & 0 & \dots & -1 & 1 & 0 & \dots & 0 \\
  0 & 0 & \dots & 0 & 1 & -1 & \dots & 0 \\
  & & \ddots & & \vdots & & \ddots & \\
  0 & 0 & \dots & 0 & 1 & 0 & \dots & -1
\end{bmatrix}.
\end{equation}
It can be seen, that $D_yD_y^T$ is independent of $y$:
\begin{equation}
\Sigma = D_yD_y^T = \begin{bmatrix}
  2 & 1 & \dots & 1 \\
  1 & 2 & \dots & 1 \\
    &   & \ddots & \\
  1 & 1 & \dots & 2
\label{eq:app-sigma}
\end{bmatrix}.
\end{equation}
Matrix $\Sigma$ and it's Cholesky decomposition, used in Genz algorithm, can be computed before training. On the other hand, the special form of $\Sigma$ largely simplifies equations for computing $\hat m_i$ and $\hat \Sigma_i$.

Cholesky decomposition of the matrix $\Sigma$ has the following form:
\begin{equation}
L = \begin{bmatrix}
  \alpha_1 & 0 & \dots & 0 \\
  \beta_1 & \alpha_2 & \dots & 0 \\
  & & \ddots & \\
  \beta_1 & \beta_2 & \dots & 0 \\
  \beta_1 & \beta_2 & \dots & \alpha_{C-1}
\end{bmatrix}.
\end{equation}
Note, that elements in each column below the diagonal are the same. Genz algorithm can be optimized to make use of this property of the matrix $L$.

\section{Backpropagation with Log-derivative Trick}

EXACT loss evaluates expectation of the form:
\begin{equation}
    \mathcal{L}(x, y, \theta) = \mathbb{E}_{s \sim \mathcal{N}(\mu, \Sigma)}\mathbbm{1}(\tilde{y}(s) = y),
\end{equation}
where $\mu = \mu(x, \theta)$ and $\Sigma = \Sigma(x, \theta)$.
While EXACT exploits properties of the multivariate normal distribution, there is a more general approach to compute the expectation and its gradients, called REINFORCE \cite{williams1992reinforce}. The expectation itself can be computed using Monte-Carlo integration:
\begin{equation}
    \mathcal{L}(x, y, \theta)  \approx \frac{1}{N}\sum\limits_{i = 1}^N \mathbbm{1}(\tilde{y}(s_i) = y),
\end{equation}
where $s_i$ are drawn from $\mathcal{N}(\mu, \Sigma)$. Gradient computation can be performed using the log-derivative trick:
\begin{equation}
    \nabla_\theta \mathcal{L}(x, y, \theta) = \mathbb{E}_{s \sim \mathcal{N}(\mu, \Sigma)}\mathbbm{1}(\tilde{y}(s) = y)\nabla\log \mathcal{N}(s;\mu, \Sigma),
\end{equation}
where $\mathcal{N}(s;\mu, \Sigma)$ is a multivariate normal PDF function. The latter expectation can also be evaluated using Monte-Carlo integration.

\begin{figure}[t]
\vskip 0.in
\centering
\includegraphics[width=\columnwidth]{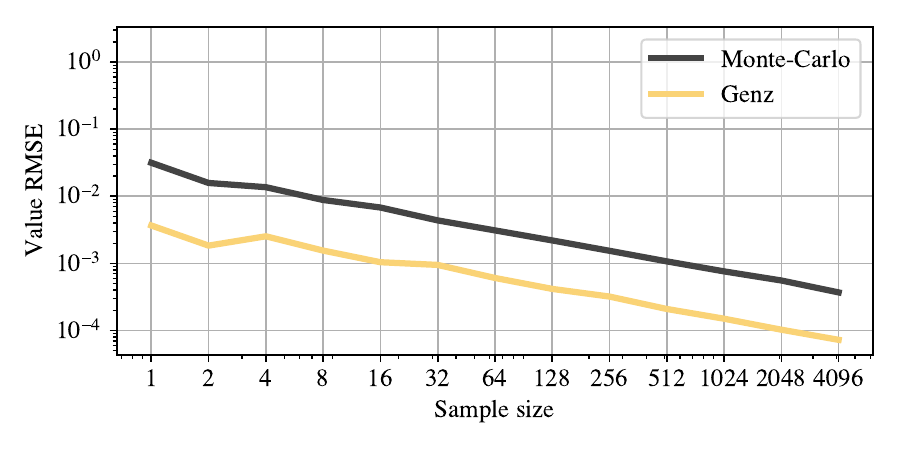}
\caption{Integral value error for Genz and Monte-Carlo algorithms with different sample sizes.}
\label{fig:mc_vs_genz_val}
\vskip -0.2in
\end{figure}

\begin{figure}[t]
\vskip 0.in
\centering
\includegraphics[width=\columnwidth]{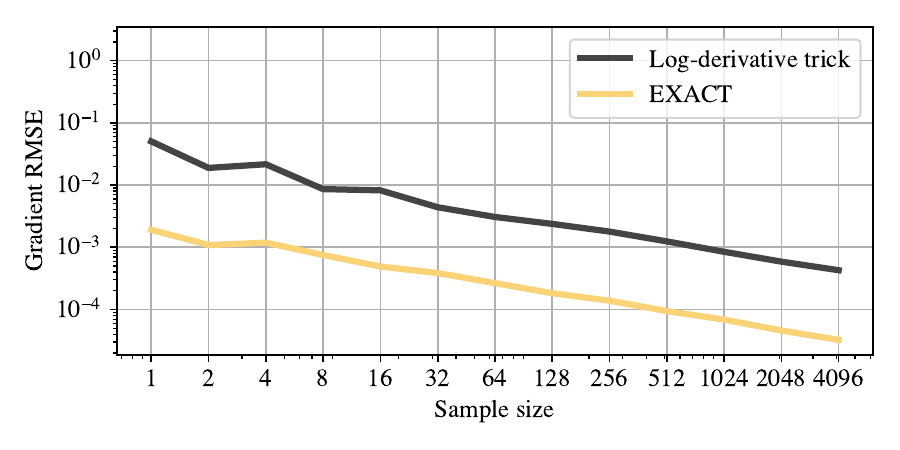}
\caption{Gradient error for EXACT and Log-derivative trick with different sample sizes.}
\label{fig:mc_vs_genz_grad}
\vskip -0in
\end{figure}

We compared the performance of Genz integration algorithm \cite{genz1992numerical} and EXACT extensions for gradient computation, with Monte-Carlo integration and log-derivative trick. For this purpose we estimated the integral from Equation \ref{eq:app-exact-comp} and its gradient from Equation (\ref{eq:app-exact-grad}). The covariance matrix $\Sigma$ was computed using Equation \ref{eq:app-sigma} and $\mu$ was set to $(1, 2, 0.5, 10, 6, -3, -4, 5, 1, 0)$. Each method was applied 1000 times with different random seeds. After that, we computed root-mean-square error (RMSE) of the integral values and ground truth value, computed with the sample size equal to 1000000.
The results are presented in Figure \ref{fig:mc_vs_genz_val} and Figure \ref{fig:mc_vs_genz_grad}. It can be seen that EXACT algorithm produces an order of magnitude smaller error than Monte-Carlo with log-derivative trick. For example, Monte-Carlo requires a sample size 256 to achieve the gradient error of EXACT with sample size 1. This study highlights the importance of the proposed EXACT computation algorithm for stable training.

\section{Toy Example Derivation}\label{app:toy}
Cross-entropy loss for the toy example has the following form:
\begin{equation}
\begin{aligned}
&\mathcal{L}_{CE}(b) = -\log \left(\frac{1}{1+\exp (-(0.25+b))}\right)- \\
&\log \left(\frac{1}{1+\exp (-b)}\right)-\log \left(\frac{1}{1+\exp (-(0.25-b))}\right).
\end{aligned}
\end{equation}
By taking derivative of the loss function w.r.t. $b$, we get
\begin{equation}
\frac{d}{db}\mathcal{L}_{CE}(b) = \frac{-3.933 e^{b}+1.284 e^{3 b}-2.568}{\left(e^{b}+1\right)\left(1.284 e^{b}+1\right)\left(1.284+e^{b}\right)}.
\end{equation}
The derivative equals zero at single point $b \approx 0.7$. As $\mathcal{L}_{CE}(b)$ is convex, it has single global minimum at $b \approx 0.7$.

The same procedure can be applied to minimize Hinge loss:
\begin{equation}
\begin{aligned}
\mathcal{L}_{Hinge}(b) = &\max (0,1+(-0.25-b))\\
+&\max (0,1-b)\\
+&\max (0,1-(0.25-b))
\end{aligned}
\end{equation}

\begin{equation}
\frac{d}{db}\mathcal{L}_{Hinge}(b) = \begin{cases}-2 & b<-\frac{3}{4} \\ -1 & -\frac{3}{4}<b<\frac{3}{4} \\ 0 & \frac{3}{4}<b<1 \\ 1 & b>1 \\ \text{N/A}& \text{otherwise} \end{cases}
\end{equation}
$\mathcal{L}_{Hinge}(b)$ reaches minimum at $0.75 \le b \le 1$.

\section{Toy Example with Learned Weight}

\begin{figure}[t]
\vskip 0.2in
\centering
\includegraphics{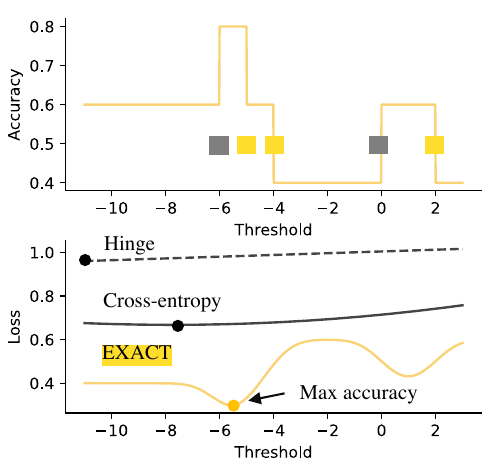}
\caption{The toy example, which demonstrates importance of accuracy optimization. The model is a linear binary classifier with weight and bias parameters. EXACT achieves 80\% accuracy, while cross-entropy and hinge loss minimization leads to 60\% accuracy (global optimum).}
\label{fig:toy2d}
\vskip -0.2in
\end{figure}

In the toy example from the main paper, we used the model with a single bias parameter, while weight was fixed to 1. Here we demonstrate another example, presented in Figure \ref{fig:toy2d}, with a trainable weight parameter. The dataset consists from 5 points in positions -6, -5, -4, 0, and 2. The labels are -1, 1, 1, -1, and 1. We minimized cross-entropy and hinge losses using grid search. Cross-entropy minimum is achieved for the threshold equal to -7.75, hinge loss is minimized at -11. Both cross-entropy and hinge loss lead to an accuracy 60\%. Gradient descent optimization of EXACT leads to the threshold -5.35 and accuracy 80\%.

\begin{table*}[t]
\centering
\begin{tabular}{lc|ccc|c}
Dataset & Split & W/O Margin & W/O Variance Scheduler & W/O Grad. Norm & EXACT \\
\hline

 adult                   & test & $84.76 \pm 0.03$      & $84.43 \pm 0.13$     & $\bf 85.27 \pm 0.02$      & $85.20 \pm 0.02$      \\
 annealing               & test & $99.00 \pm 0.00$      & $92.20 \pm 1.17$     & $98.00 \pm 0.63$      & $\bf 100.00 \pm 0.00$ \\
 audit-risk              & test & $94.74 \pm 0.48$      & $\bf 97.31 \pm 0.48$     & $93.72 \pm 0.48$      & $96.92 \pm 0.48$      \\
 balance-scale           & test & $\bf 92.00 \pm 0.00$  & $\bf 92.00 \pm 0.00$ & $\bf 92.00 \pm 0.00$  & $\bf 92.00 \pm 0.00$  \\
 breast-cancer-wisconsin & test & $95.61 \pm 0.00$      & $96.32 \pm 0.66$     & $\bf 97.54 \pm 0.66$  & $96.49 \pm 0.00$  \\
 car                     & test & $91.04 \pm 0.00$      & $93.58 \pm 0.22$ & $94.68 \pm 0.29$  & $\bf 94.80 \pm 0.00$  \\
 cylinder-bands          & test & $73.15 \pm 0.59$      & $69.81 \pm 1.99$     & $76.48 \pm 0.94$  & $\bf 77.22 \pm 0.45$  \\
 dry-bean                & test & $92.41 \pm 0.05$      & $\bf 92.97 \pm 0.15$     & $92.56 \pm 0.22$      & $92.79 \pm 0.08$      \\
 mice-protein            & test & $98.15 \pm 0.00$      & $97.13 \pm 0.68$     & $\bf 98.70 \pm 0.45$      & $98.33 \pm 0.56$      \\
 wine                    & test & $\bf 100.00 \pm 0.00$ & $99.44 \pm 1.11$     & $\bf 100.00 \pm 0.00$ & $\bf 100.00 \pm 0.00$ \\

\end{tabular}

\caption{Ablation studies for different EXACT variants.}
\label{tab:ablations}
\end{table*}

\begin{table*}[t]
\centering
\begin{tabular}{ll|cc}
Dataset & Split & Cross-entropy with gradient normalizer & Cross-entropy \\
\hline
adult                   & test & $85.27 \pm 0.01$ & $\bf 85.28 \pm 0.01$ \\
annealing               & test & $\bf 100.00 \pm 0.00$ & $\bf 100.00 \pm 0.00$ \\
audit-risk              & test & $\bf 98.08 \pm 0.00$ & $96.28 \pm 0.48$ \\
balance-scale           & test & $\bf 92.00 \pm 0.00$ & $\bf 92.00 \pm 0.00$ \\
breast-cancer-wisconsin & test & $95.61 \pm 0.00$ & $\bf 95.96 \pm 0.43$ \\
car                     & test & $\bf 92.20 \pm 0.00$ & $91.91 \pm 0.00$ \\
cylinder-bands          & test & $\bf 72.78 \pm 0.74$ & $\bf 72.78 \pm 0.45$ \\
dry-bean                & test & $92.81 \pm 0.01$ & $\bf 92.83 \pm 0.01$ \\
mice-protein            & test & $98.61 \pm 0.00$ & $\bf 98.89 \pm 0.23$ \\
wine                    & test & $\bf 100.00 \pm 0.00$ & $\bf 100.00 \pm 0.00$ \\
\hline
Total best & & 6 & 8 \\

\end{tabular}

\caption{Comparison of cross-entropy loss with and without gradient normalizer on tabular datasets.}
\label{tab:ablation-gnorm-xent}
\end{table*}

\begin{table*}[t]
\centering
\begin{tabular}{ll|cc}
Dataset & Split & Hinge loss with gradient normalizer & Hinge loss \\
\hline
adult                    & test & $85.23 \pm 0.03$ & $\bf 85.26 \pm 0.01$ \\
annealing                & test & $\bf 100.00 \pm 0.00$ & $\bf 100.00 \pm 0.00$ \\
audit-risk               & test & $\bf 98.72 \pm 0.00$ & $97.05 \pm 0.65$ \\
balance-scale            & test & $\bf 92.00 \pm 0.00$ & $\bf 92.00 \pm 0.00$ \\
breast-cancer-wisconsin  & test & $\bf 97.37 \pm 0.00$ & $96.49 \pm 0.00$ \\
car                      & test & $90.17 \pm 1.92$ & $\bf 92.77 \pm 0.00$ \\
cylinder-bands           & test & $73.15 \pm 0.00$ & $\bf 74.44 \pm 0.45$ \\
dry-bean                 & test & $92.75 \pm 0.04$ & $\bf 92.93 \pm 0.03$ \\
mice-protein             & test & $\bf 98.98 \pm 0.19$ & $97.69 \pm 0.00$ \\
wine                     & test & $\bf 100.00 \pm 0.00$ & $\bf 100.00 \pm 0.00$ \\
\hline
Total best & & 6 & 7 \\

\end{tabular}

\caption{Comparison of hinge loss with and without gradient normalizer on tabular datasets.}
\label{tab:ablation-gnorm-hinge}
\end{table*}

\section{Ablation Studies}
In this section we compare methods with and without \mbox{EXACT} improvements on UCI datasets. First, we compare EXACT variants without margin, without variance scheduler, and without gradient normalizer with original EXACT method. Results are presented in Table \ref{tab:ablations}. According to these results, all improvements generally lead to higher test-time accuracy.

We also evaluated the effect of gradient normalizer in the context of cross-entropy and hinge loss. Results are presented in Table \ref{tab:ablation-gnorm-xent} and Table \ref{tab:ablation-gnorm-hinge}. According to these results, gradient normalizer has minor effect on training with baseline loss functions.

\section{Hyperparameters}

Hyperparameter ranges and values found by 50 iterations of random search for each loss function for image classification datasets are listed in Table \ref{tab:hopt-vision}. Ranges and hyperparameters for tabular datasets are presented in Table \ref{tab:hopt-tabular}.
\begin{table*}[t]
\centering
\begin{tabular}{l|l|cccc}
Method & Dataset & SGD/ASAM Learning rate & Adam Learning rate & Gradient clipping & Margin \\
\hline
Range & & 0.001 -- 1 & 0.001 -- 1 & 0.01 -- 10 & 0 -- 10 \\
\hline
\multirow{3}{*}{Cross-entropy}
& MNIST & 0.5 & 0.001 & 10.0 & N/A \\
& SVHN & 1.0 & 0.001 & 0.1 & N/A \\
& CIFAR-10 & 0.1 & 0.001 & 10.0 & N/A \\
& CIFAR-100 & 0.5 & 0.001 & 1.0 & N/A \\
\hline
\multirow{3}{*}{Hinge}
& MNIST & 0.05 & 0.001 & 1.0 & 5.0 \\
& SVHN & 0.5 & 0.001 & 0.1 & 10.0 \\
& CIFAR-10 & 0.1 & 0.001 & 10.0 & 1.0 \\
& CIFAR-100 & 1.0 & 0.001 & 10.0 & 0.1 \\
\hline
\multirow{3}{*}{EXACT}
& MNIST & 0.5 & 0.001 & N/A & 0.5 \\
& SVHN & 1.0 & 0.001 & N/A & 10.0 \\
& CIFAR-10 & 1.0 & 0.001 & N/A & 0.5 \\
& CIFAR-100 & 1.0 & 0.001 & N/A & 5.0 \\
\end{tabular}

\caption{Hyperparameters for vision classification datasets.}
\label{tab:hopt-vision}
\end{table*}

\begin{table*}[t]
\centering
\begin{tabular}{l|l|cccc}
Method & Dataset & Learning rate & Gradient clipping & Margin & Regularization \\
\hline
Range & & 0.01 -- 5 & 0.01 -- 10 & 0 -- 10 & 0 -- 1 \\
\hline
\multirow{10}{*}{Sklearn} & audit-risk & N/A & N/A & N/A & 1 \\
 & adult & N/A & N/A & N/A & 10 \\
 & annealing & N/A & N/A & N/A & 0 \\
 & balance-scale & N/A & N/A & N/A & 0 \\
 & breast-cancer-wisconsin & N/A & N/A & N/A & 0.1 \\
 & car & N/A & N/A & N/A & 0.1 \\
 & cylinder-bands & N/A & N/A & N/A & 1 \\
 & dry-bean & N/A & N/A & N/A & 0 \\
 & mice-protein & N/A & N/A & N/A & 0 \\
 & wine & N/A & N/A & N/A & 0 \\
\hline
\multirow{10}{*}{Cross-entropy} & audit-risk & 0.01 & 0.1 & N/A & 0 \\
 & adult & 1 & 0.01 & N/A & 0 \\
 & annealing & 1 & 10.0 & N/A & 0 \\
 & balance-scale & 1 & 10.0 & N/A & 0 \\
 & breast-cancer-wisconsin & 0.01 & 0.01 & N/A & 0 \\
 & car & 1 & 10.0 & N/A & 0 \\
 & cylinder-bands & 0.5 & 0.01 & N/A & 0 \\
 & dry-bean & 0.5 & 0.1 & N/A & 0 \\
 & mice-protein & 0.05 & 10.0 & N/A & 0 \\
 & wine & 1 & 10.0 & N/A & 0 \\
\hline
\multirow{10}{*}{Hinge} & audit-risk & 0.1 & 0.1 & 0 & 0 \\
 & adult & 0.1 & 0.1 & 0.1 & 0 \\
 & annealing & 1 & 10.0 & 0.5 & 0 \\
 & balance-scale & 1 & 10.0 & 0.5 & 0 \\
 & breast-cancer-wisconsin & 0.01 & 1.0 & 5 & 0 \\
 & car & 1 & 0.01 & 0.5 & 0 \\
 & cylinder-bands & 0.5 & 0.01 & 2 & 0 \\
 & dry-bean & 1 & 0.01 & 0.5 & 0\\
 & mice-protein & 0.01 & 0.1 & 1 & 0 \\
 & wine & 1 & 10.0 & 0.5 & 0 \\
\hline
\multirow{10}{*}{EXACT} & audit-risk & 0.05 & N/A & 0.05 & 0 \\
 & adult & 5 & N/A & 0.1 & 0 \\
 & annealing & 5 & N/A & 0.1 & 0 \\
 & balance-scale & 1 & N/A & 0.01 & 0 \\
 & breast-cancer-wisconsin & 5 & N/A & 5 & 0 \\
 & car & 0.05 & N/A & 0.05 & 0 \\
 & cylinder-bands & 5 & N/A & 0.1 & 0 \\
 & dry-bean & 5 & N/A & 0.1 & 0 \\
 & mice-protein & 0.1 & N/A & 1 & 0 \\
 & wine & 0.05 & N/A & 5 & 0 \\
\hline
\end{tabular}
\caption{Hyperparameters for tabular classification datasets.}
\label{tab:hopt-tabular}
\end{table*}

\end{document}